\RequirePackage{lineno}
\documentclass[aapm,graphicx]{revtex4-1}
\usepackage[utf8]{inputenc}
\usepackage{color,soul}
\usepackage{graphicx}
\graphicspath{}
\usepackage[colorlinks=true,linkcolor=blue]{hyperref}
\usepackage{amsmath}
\usepackage{float}

\usepackage{soul}
\usepackage{mathtools}

\usepackage{colortbl}
\DeclarePairedDelimiter\ceil{\lceil}{\rceil}

\usepackage{soul}

\usepackage{hyperref}

\makeatletter
\newcommand{\printfnsymbol}[1]{%
  \textsuperscript{\@fnsymbol{#1}}%
}
\makeatother

\usepackage[symbol]{footmisc}
\usepackage{lineno}
\usepackage{fancyhdr}
\begin{document}
\title{Targeted transfer learning to improve performance in small medical physics datasets}

\author{Miguel Romero BSc$^{1*}$\footnote[0]{* Equal contribution authors.}, Yannet Interian PhD$^{1*+}$\footnote[0]{$^+$ Corresponding author: 101 Howard st San Francisco, CA 94105 (yinterian@usfca.edu).}, Timothy Solberg PhD$^{2}$ and Gilmer Valdes PhD$^{2}$ }

\address{$^1$Master of Science in Data Science, University of San Francisco, San Francisco, CA, 94105, USA;
$^2$Department of Radiation Oncology, University of California San Francisco, San Francisco, CA, 94158, USA
}

\date{\today}

\begin{abstract}

\textbf{Purpose:} To perform an in depth evaluation of current state of the art techniques in training neural networks to identify appropriate approaches in small datasets.

\textbf{Method:} 112,120 frontal-view x-ray images from the NIH ChestXray14 dataset were used in our analysis. Two tasks were studied: unbalanced multi-label classification of 14 diseases, and binary classification of pneumonia vs non-pneumonia. All datasets were randomly split into training, validation and testing (70\%, 10\%, 20\%). Two popular convolution neural networks (CNN), DensNet121 and ResNet50 were trained using PyTorch. We performed several experiments to test: 1) whether transfer learning using pre-trained networks on ImageNet are of value to medical imaging / physics tasks (e.g., predicting toxicity from radiographic images after training on images from the internet), 2) whether using pre-trained networks trained on problems that are similar to the target task helps transfer learning (e.g., using x-ray pre-trained networks for x-ray target tasks), 3) whether freeze deep layers or change all weights provides an optimal transfer learning strategy, 4) the best strategy for the learning rate policy and, 5) what quantity of data is needed in order to appropriately deploy these various strategies ($N=50$ to $N=77,880$). 

\textbf{Results:} In the multi-label problem, DensNet121 needed at least 1600 patients to be comparable to, and 10,000 to outperform, radiomics-based logistic regression. In classifying pneumonia vs. non-pneumonia, both CNN and radiomics-based methods performed poorly when $N < 2000$. For small datasets ($< 2000$), however, a significant boost in performance ($> 15\%$ increase on AUC) comes from a good selection of the transfer learning dataset, dropout, cycling learning rate and freezing and unfreezing of deep layers as training progresses. In contrast, if sufficient data is available ($>35000$), little or no tweaking is needed to obtain impressive performance. While transfer learning using x-ray images from other anatomical sites improves performance, we also observed a similar boost by using pre-trained networks from ImageNet. Having source images from the same anatomical site, however, outperforms every other methodology, by up to 15\%. In this case, DL models can be trained with as little as $N=50$. 

\textbf{Conclusions:} While training DL models in small datasets ($N < 2000$) is challenging, no tweaking is necessary for bigger datasets ($N > 35000$). Using transfer learning with images from the same anatomical site can yield remarkable performance in new tasks with as few as N=50. Surprisingly, we did not find any advantage for using images from other anatomical sites over networks that have been trained using ImageNet.  This indicates that features learned may not be as general as currently believed, and performance decays rapidly even by just changing the anatomical site of the images. 

\end{abstract}

\maketitle

\section{Introduction}

The use of machine learning in medical imaging, radiation theranostics and medical physics applications has created tremendous opportunity with research that encompasses: quality assurance \cite{valdes2016mathematical, valdes2017imrt, interian2018deep, valdes2015use, zhu2011planning,carlson2016machine}, outcome prediction \cite{valdes2016using, valdes2018salvage, luna2019predicting, hara2018clinical, el2009predicting, zhen2017deep, ibragimov2018development}, segmentation \cite{ibragimov2017segmentation,ibragimov2017combining, qin2018superpixel, kearney2018unsupervised} or dosimetric prediction \cite{valdes2017clinical, shiraishi2016knowledge, nguyen2017dose}. Specifically, due to successes in other fields, deep learning techniques are attracting particular attention for image classification tasks. The current popularity of deep learning algorithms began in 2012 when Krizhevsky et al won the Imagenet Large-Scale Visual Recognition Challenge (ILSVRC) using  a convolutional neural network (CNN) \cite{krizhevsky2012Imagenet}. Their CNN achived a top-5 test error rate of 15.3\%, compared to 26.2\% achieved by the second-best entry
using. Many other applications quickly followed (e.g. machine translation) with similar results. Not to be left behind, many applications of deep learning algorithms in radiation oncology, medical imaging and medical physics are proliferating in the literature 
\cite{interian2018deep, ibragimov2018development, ibragimov2017segmentation, qin2018superpixel, kearney2018unsupervised, nguyen2017dose}. Datasets in radiation oncology and medical physics, however, tend to be small in size. They are generally in the hundreds and rarely in the thousands or millions of images typical of other industries. Further, the extremely large number of parameters, and dozens of hyper parameters that need to be tuned to train neural network raises the question of whether deep learning algorithms are truly appropriate when images are analyzed in our field. Within the machine learning literature, many papers advise against the use of deep learning when limited data is available. “In the future, based on the observations from this study, the development of similar high-performing algorithms for medical imaging using deep learning has two prerequisites. First, there must be collection of a large developmental set with tens of thousands of abnormal cases…."  \cite{Gulshan}. 

In order to ameliorate the effect of small datasets, the use of transfer learning is ubiquitous  \cite{interian2018deep, ibragimov2018development, ibragimov2017segmentation, qin2018superpixel, kearney2018unsupervised, nguyen2017dose}. In transfer learning, features from a network trained in a “source” task are transferred to a second “target” network with the hope that the features learned will help in the new problem. Transfer learning is not, however, a magic bullet. The extent of training that can be performed in the target task depends on the size of the target dataset, since the algorithms can easily overfit in small datasets \cite{deepNN}. It has been shown that the performance of transferring features decreases the more dissimilar the source and target are \cite{deepNN}. This directly points to a possible problem, since in the medical imaging literature most transfer learning is performed using a network trained on Imagenet (i.e., predicting classes such as cats or dogs from color images) \cite{interian2018deep, ibragimov2018development, ibragimov2017segmentation, qin2018superpixel, kearney2018unsupervised, nguyen2017dose}. Additionally, and in contrast to popular belief, several authors have recently indicated that transfer learning might not be as general as previously thought. They demonstrated that transfer learning does not always result in better performance, that features are not easily transferred to other tasks (e.g., from Imagenet to x-rays) or its value is more related to providing a good initialization for the parameters of the networks as opposed to reusing the features \cite{MedicalImages,rethinking,betterImagenet}. However, this work was performed in datasets with tenth of thousands of data points where transfer learning is not as critical.

In the present article, therefore, we investigate: 1) whether pre-trained networks on Imagenet are of use to medical imaging / physics tasks (e.g., predicting toxicity from radiographic images after training on images from the internet); 2) whether using pre-trained networks trained on problems that are similar to the target task helps transfer learning (e.g., using x-ray pre-trained networks for x-ray target tasks); 3) what the best training strategies are when there is a small amount of data; and 4) what quantity of data is needed in order to appropriately deploy the various strategies. We address these questions by investigating the performance of deep learning algorithms as a function of the size of the training dataset, the methods used to train, and the transfer learning strategies. We also compare the performance against baseline methods that use radiomic features \cite{rad1} and statistical learning algorithms. 

Further we present an in depth evaluation of the effect of data size on the performance of CNN for medical physics problems with subsequent recommendations, we introduce a novel and effective training method in small datasets, we propose an alternative use of transfer learning in medical imaging / physics, and we demonstrate experimentally the most effective techniques for training and transfer learning. Finally, to guarantee scientific reproducibility, we are releasing our code to facilitate the replication of our results by other investigators. All datasets used are publicly available.

\section{Methods and Materials}\label{sect: methods and materials}

\subsection{Datasets}

In transfer learning there are two tasks: the “source” task, generally a large dataset on which pre-training is performed (e.g., Imagenet, which contains 1.2 million images with 1000 categories), and the “target” task of interest. In this work, source refers to the dataset or task with which the network is first trained, and target refers to the dataset or task with which the network is fine-tuned. The following describes the datasets used in this study and how they are used.

\begingroup
\setlength{\tabcolsep}{15pt} 
\renewcommand{\arraystretch}{1.5} 
\begin{table}[h!]
\centering
\begin{tabular}{l l  l}
\hline
Dataset & Partition & \# Positive observations \\
\hline
Emphysema & training & 1272\\
          & validation & 151\\
          & testing & 1093\\
Hernia    & training & 121\\
           & validation & 20\\
           & testing & 86\\
Pneumonia & training & 779\\
          & validation & 55\\
          & testing & 97\\
\hline
\end{tabular}
\caption{\label{table: positive cases} Number of positive cases in the target binary training, validation and testing datasets.} 
\end{table}
\endgroup

\subsubsection{Source Datasets}

\begin{itemize}

    \item\underline{MURA dataset}: Consists of $14,863$ musculoskeletal radiographs studies from $12,173$ patients, with a total of $40,561$ multi-view radio-graphic images. Each belongs to one of seven standard upper extremity radio-graphic study types: elbow, finger, forearm, hand, humerus, shoulder, and wrist. The labels are 1 and 0 for normal and abnormal respectively \cite{MURA}.
    \item \underline{CheXpert dataset}: Consists of $224,316$ chest radio-graphs from $65,240$ patients labeled for 14 diseases as negative, positive or uncertain. From those classes, 5 of them were analyzed: Atelectasis, Cardiomegaly, Consolidation, Edema and Pleural Effusion. One or zero were assigned to the uncertain labels of each class according to what yield the best result reported in \cite{chexpert}. All of the five disease are also considered in Chest X-ray 14 \cite{chexpert}.
    \item \underline{Chest X-ray13 dataset}: Constructed from the Chest X-ray14 with a sample of approximately 75 thousand negative cases of Emphysema. The labels in this dataset correspond to the 14 original diseases but excluding Emphysema (13 diseases). This dataset has no intersection with the Emphysema dataset (described below) \cite{WangX}.
   
\end{itemize}

\subsubsection{Target Datasets}

\begin{itemize}
     \item \underline{Chest X-ray14 dataset}: The original dataset contains $112,120$ frontal-view chest X-ray images and 14 disease labels per image \cite{WangX}.
     \item \underline{Pneumonia dataset}: Contains image-label pairs where the images are unmodified from ChestX-ray14 and the labels are 1 or 0 corresponding to the ground truth for Pneumonia \cite{WangX}.
     \item \underline{Emphysema dataset}: Contains image-label pairs where the images are unmodified from ChestX-ray14 and the labels are 1 or 0 corresponding to the ground truth for Emphysema \cite{WangX}.
     \item \underline{Hernia dataset}: Contains image-label pairs where the images are unmodified from ChestX-ray14 and the labels are 1 or 0 corresponding to the ground truth for Hernia \cite{WangX}.
\end{itemize}

With medical images and disease detection, it is often the case that the number of positive cases is several orders of magnitude smaller than the number of negative cases. The datasets used here are no exception. Thus the validation and testing datasets of binary tasks were balanced by sampling negative observations in such a way that the number of unique positive observations was maximized. This was performed once for each experiment. The performance of each model trained in a different number of training observations is reported using the same testing dataset for each specific experiment. For training, sampling is conducted once for each experiment and number of training samples. For the target binary datasets, the specific number of available positive observations for training, validation and test is shown in Table \ref{table: positive cases}.

\subsection{Radiomic methods}
 
In order to establish baseline performance for each task, radiomic features and statistical learning algorithms were used. The traditional approaches were constructed using 79 numerical radiomic features extracted from the images: grey level co-occurrence matrix (GLCM) features, with 1; 2; 4; 6 pixel offsets and 0; 45 and 90-degree angles (72 features/image), grey image mean, standard deviation and second, third, fourth and fifth moment (6 features/image), and the image entropy (1 feature/image) \cite{rad1, rad2, rad3}. This set of features cannot be considered exhaustive but serves as reference to benchmark the traditional approaches across the spectrum of available training data since they include well established GLCM and statistical features. The classifiers were built using logistic regression with L1 regularization (Lasso), L2 regularization (Ridge), Elastic Net (linearly combines the L1 and L2 penalties of the lasso and ridge), and random forest. These models were tuned using cross validation and random search for the different hyper parameters. The best traditional model was reported. 

\subsection{Data augmentation and testing time augmentation for deep learning}

During training, all the images were initially re-sized to a width of 250, preserving the most-common aspect ratio in the original dataset. Three standard data augmentation techniques were used throughout all the experiments: 1) Random flip, with a probability $0.5$ images are reflected over the central vertical line; 2) random rotation, in which images are randomly rotated between +/-10 degrees, with uniform probability; and 3) random cropping, in which images are randomly cropped to 236x236. Testing time augmentation (TTA), which has been reported to offer better results \cite{TTA} was used. TTA is an application of data augmentation to the test dataset. Data augmentation is applied to the test set by creating 1) four different copies of an image and the original; 2) predicting on all five images; and 3) averaging the output to obtain the final prediction.

\subsection{Training methods for deep learning}

\begin{figure}[t]
\includegraphics[width=0.9\textwidth]{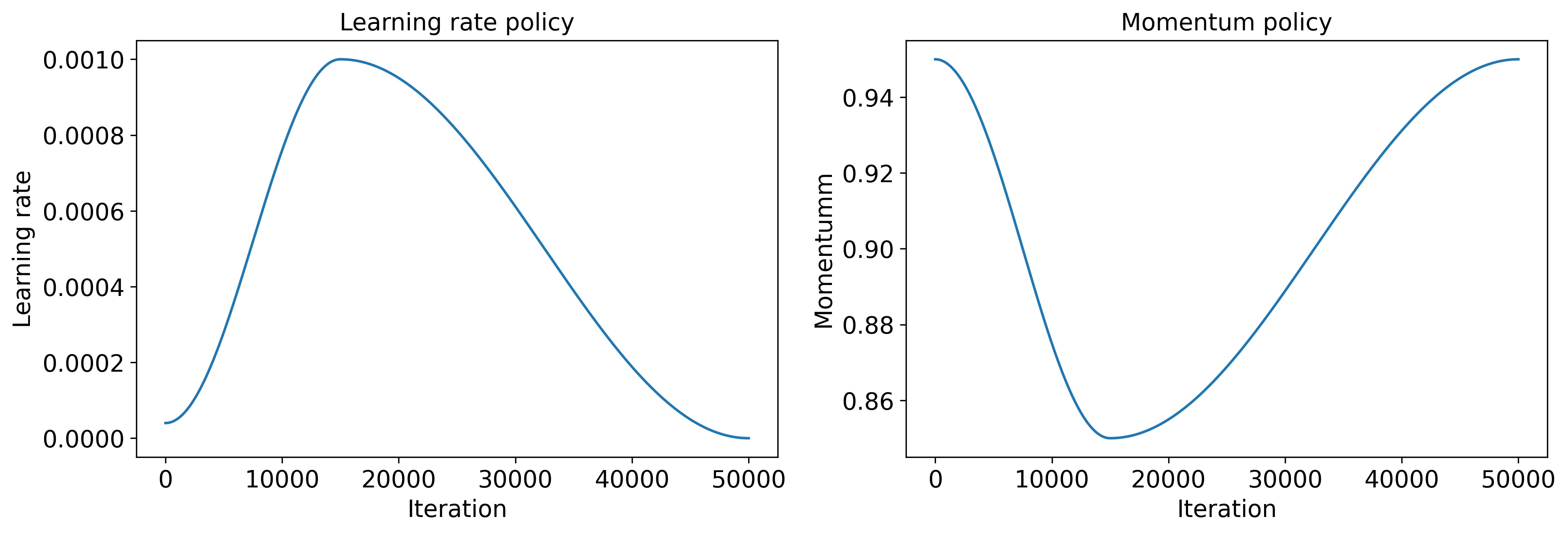}
\caption{Illustration of a one-cycle learning rate and momentum policies with cosine annealing for 500 mini-batch iterations.}
\label{fig: one-cycle}
\end{figure}

\begin{figure} 
\centering
\includegraphics[width=.5\textwidth]{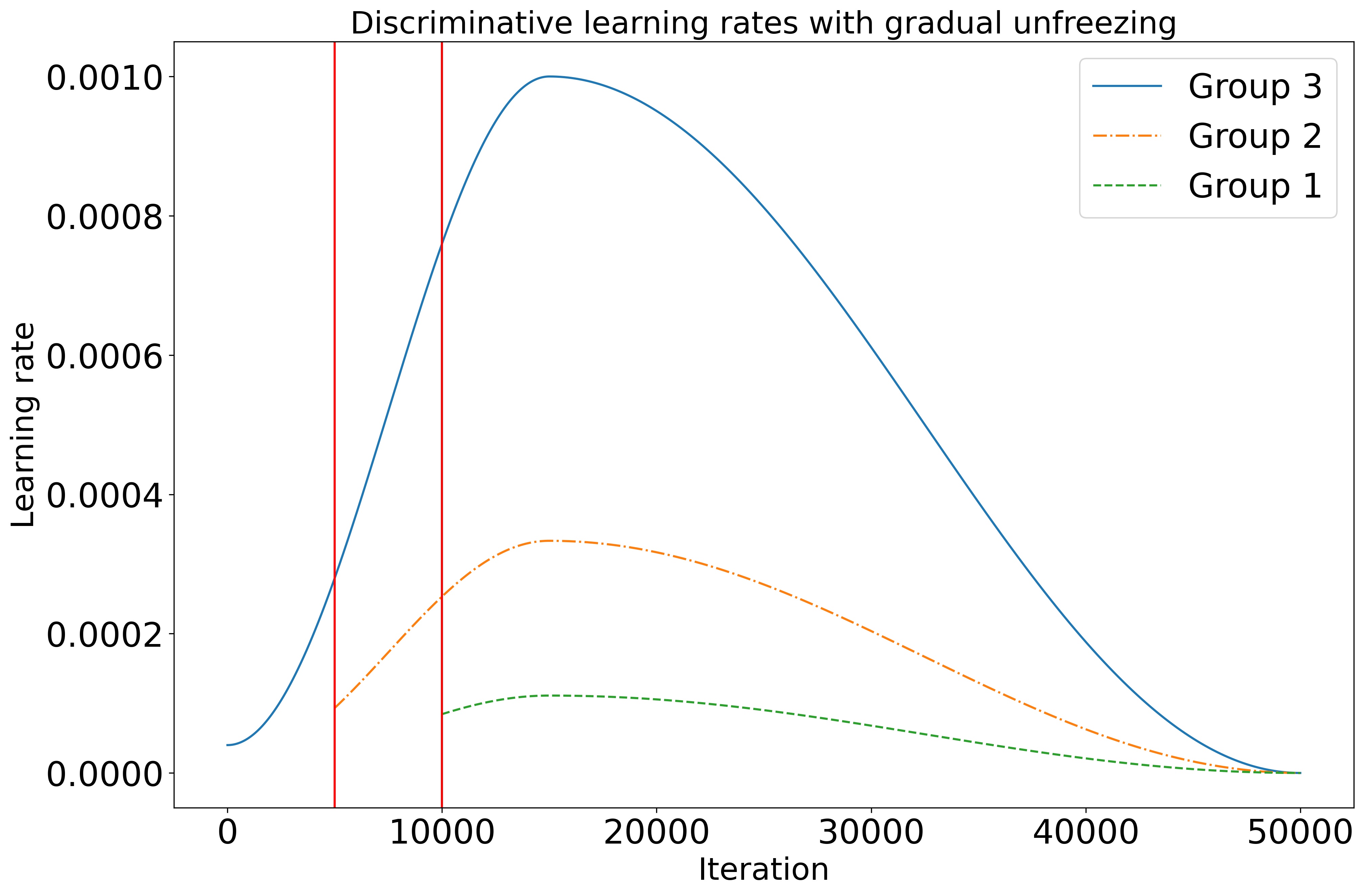}
\caption{Illustration of our implementation of one-cycle learning rate policy with 3 layer groups discriminative learning rates and gradual unfreezing. The group of layers closer to the image is unfrozen at the second red line (starting from the left). The second group of layers closer to the image is unfrozen at the first red line (starting from the left).}
\label{fig: one-cycle-grad-unfreezing-diff-lr}
\end{figure}

\begin{figure}
\centering
\includegraphics[width=1.0\textwidth]{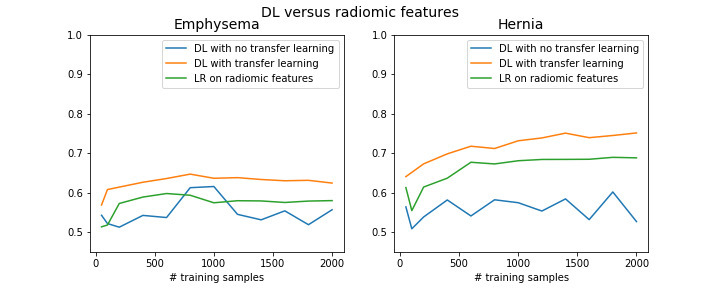}
\caption{Test AUC for the Hernia the Emphysema tasks using: deep learning without transfer learning; deep learning with transfer learning; logistic regression with radiomic features. All as a function of training dataset size.}
\label{figure: all}
\end{figure}

Learning rate is one of the most important parameters to tune when training a neural network \cite{practicalrecommendations}. In this work, we use a learning rate finder \cite{cyclical-lr} to select the largest value that can be used as a learning rate $max\_lr$. The algorithm starts by training the model with a small learning rate and increases it at each training iteration up to a maximum learning value. As the learning rate starts increasing the loss will eventually increase. The maximum learning rate, $max\_lr$,  is chosen to be the point what which the loss is minimized. 

Two training methods were considered: regular training and, a novel, one-cycle training \cite{Ng, one-cycle}. Here is a description of the two methods:

\begin{itemize}
    \item \underline{Regular training}: regular training is the most common way to train CNN. Start with an initial learning rate ($max\_lr$) and decay by a factor of 10 each time the validation loss plateaus after an epoch \cite{Ng}.

    \item \underline{One-cycle training}: in one-cycle training, the learning rate is first increased for $30\%$ of the iterations and then decreased according to the following schedule \cite{fastai}. The momentum policy follows the opposite pattern, that is, momentum is first decreased and then increased. A cosine annealing formulation is described in equation  \ref{eq: cos-annealing} where $lr_{start}$ and $lr_{end}$ are the starting and the end learning rate of a specific segment and $i$ and $T$ are the current iteration and the total number of iterations respectively. The training ends when the maximum number of epochs is reached. Figure \ref{fig: one-cycle} provides a graphical description \cite{one-cycle}.
    \begin{equation}\label{eq: cos-annealing}
        f(i)_{lr_{start}, lr_{end}} = lr_{end} + \frac{lr_{start} - lr_{end}}{2}\big( 1+ \cos\left(i\frac{\pi}{T}\right)\big)
    \end{equation}
    
    To get the full learning rate policy we use this formula:
$$
    lr(i)= 
\begin{cases}
    f(i)_{max\_lr/25, max\_lr} & i < cut\\
    f(i-cut)_{max\_lr, max\_lr/25000} & \text{otherwise}
\end{cases}
$$
Where $cut = \ceil*{0.3 \cdot max\_iter}$ and $max\_iter$ is the total number of iterations. A similar formula is used to get the momentum policy.
\end{itemize}

In both type of trainings we pick the model with the lowest validation loss.

\subsection{Transfer learning methods}

In computer vision applications, deep learning models are rarely trained from scratch, but instead transfer learning is used. In the majority of cases, models are fine-tuned using models that have been trained on Imagenet. In order to use a CNN pre-trained from Imagenet, the last fully-connected layer is removed and replaced with a randomly initialized layer of the appropriated size. When training with small datasets, it is not uncommon to keep some of the earlier layers fixed (frozen) to avoid overfitting.

The following transfer learning methods were considered in this paper: CNN as feature extractor, fine tuning all parameters simultaneously, and discriminative learning rates with gradual unfreezing.

\begin{itemize}
    \item \underline{CNN as feature extractor}: All but the last feed-forward layer(s) of the network are frozen. The only weights that are trained are those in the last layers \cite{Karpathy}.
    \item \underline{Fine tune all CNN simultaneously}: None of the weights are frozen. The pretrained network is used as a starting point \cite{Karpathy}. 
     \item \underline{Discriminative learning rates with gradual unfreezing}: This method relies on the assumption that the first layers of a network typically learn to general features (e.g., lines, circles, colors, etc). Thus, the weights in those layers should be changed less than the weights of the downstream layers which are more specialized in the target task. To do so, the network is split into three consecutive groups of layers. The learning rate then discriminates as a function of the layer, with the first group having the smallest learning rate and the last group having the highest learning rate \cite{fastai, ULMFiT}. Using gradual unfreezing, the first and second groups will be trained for fewer iterations than the third group. This was implemented as follows:
    \begin{enumerate}
        \item Generate a learning rate schedule for Group 3 using a one-cycle policy.
        \item The learning rate schedule for for Group 2 is generated by dividing by 1/3 the schedule for Group 3. Similarly, for Group 1 we divide by 1/9 the schedule of Group 3.
        \item Start by freezing all layers in Group 1 and Group 2. 
        \item After 10\% of the total iterations unfreeze Group 2.  
        \item After 20\% of the total iterations unfreeze Group 3. 
    \end{enumerate}
     
    A representation of the implementation of this one-cycle learning rate policy with discriminative learning rates and gradual unfreezing with three groups of layers is shown in Figure \ref{fig: one-cycle-grad-unfreezing-diff-lr}.
     
\end{itemize}

\subsection{Reported metric and Experimental Setup}

The area under the receiver operating characteristic curve (AUC) and the average AUC of each label is reported in the binary and multi-label tasks respectively. To evaluate best practices when using deep learning in small datasets, a stage-wise strategy was adopted. Multiple options were compared in series, one at the time, and always choosing the best to perform the next set of experiments. For instance, when evaluating transfer learning, the training method that yielded the best result in the previous stage was used. The following order was followed: training method, transfer learning method and transfer learning type of source dataset. To objectively compare techniques and methods, for each stage and target task, the validation and test data sets were held constant. The different methods and options were then evaluated for a collection of varying number of training samples, ranging from 50 to 2000. Because model training and evaluation for each methodology and number training data points is highly time consuming tuning of hyperparameters was performed with 2000 training samples, with the resulting hyperparameters applied to the rest of the training samples. In all cases, early stopping was used to prevent over-fitting. The validation loss is reported after every epoch, and the final model corresponds to the one that yields the lowest validation loss.

\begin{figure}
\centering
\includegraphics[width=0.9\textwidth]{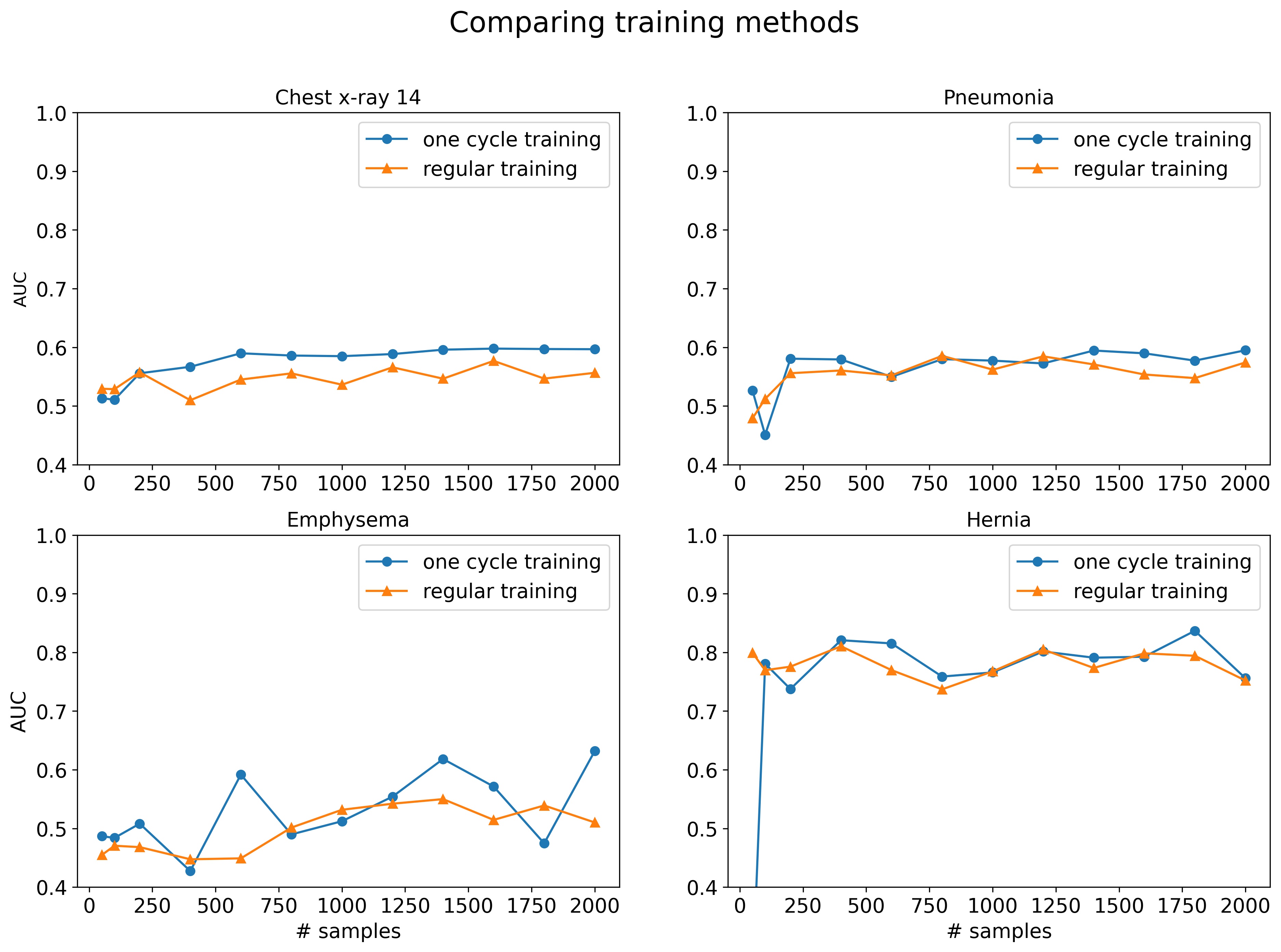}
\caption{Comparison of different training methods as a function of training dataset size. The tasks are, from left to right, top to bottom: Chest X-ray 14, Pneumonia, Emphysema, and Hernia.}
\label{figure: training methods}
\end{figure}

 \begin{figure}
\centering
\includegraphics[width=0.9\textwidth]{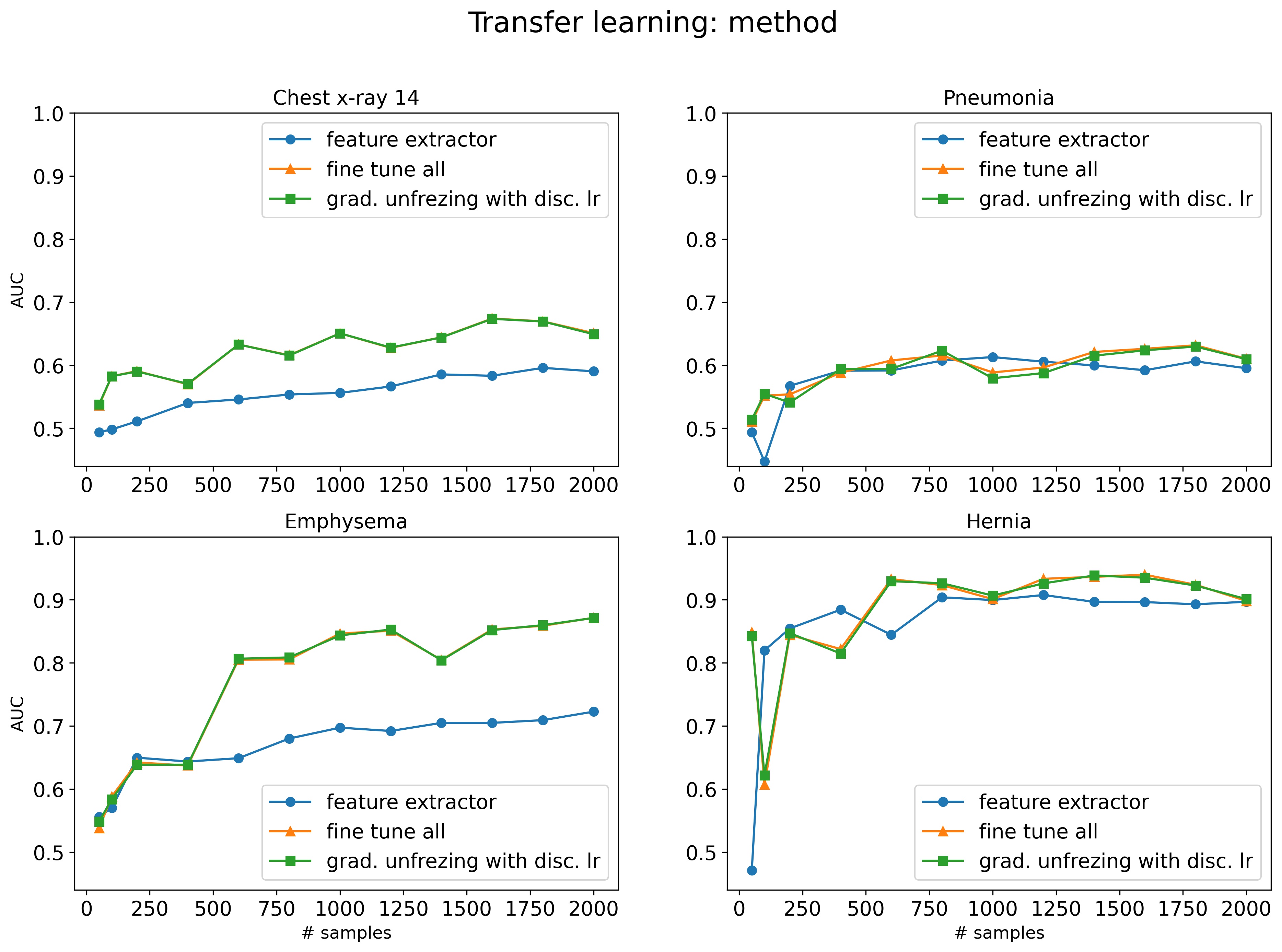}
\caption{Comparison of three different transfer learning methods, using DenseNet121, as a function of training dataset size. The tasks are, from left to right, top to bottom: Chest X-ray 14, Pneumonia, Emphysema, and Hernia.}
\label{figure: tl methods}
\end{figure}

\begin{figure}
\centering
\includegraphics[width=0.9\textwidth]{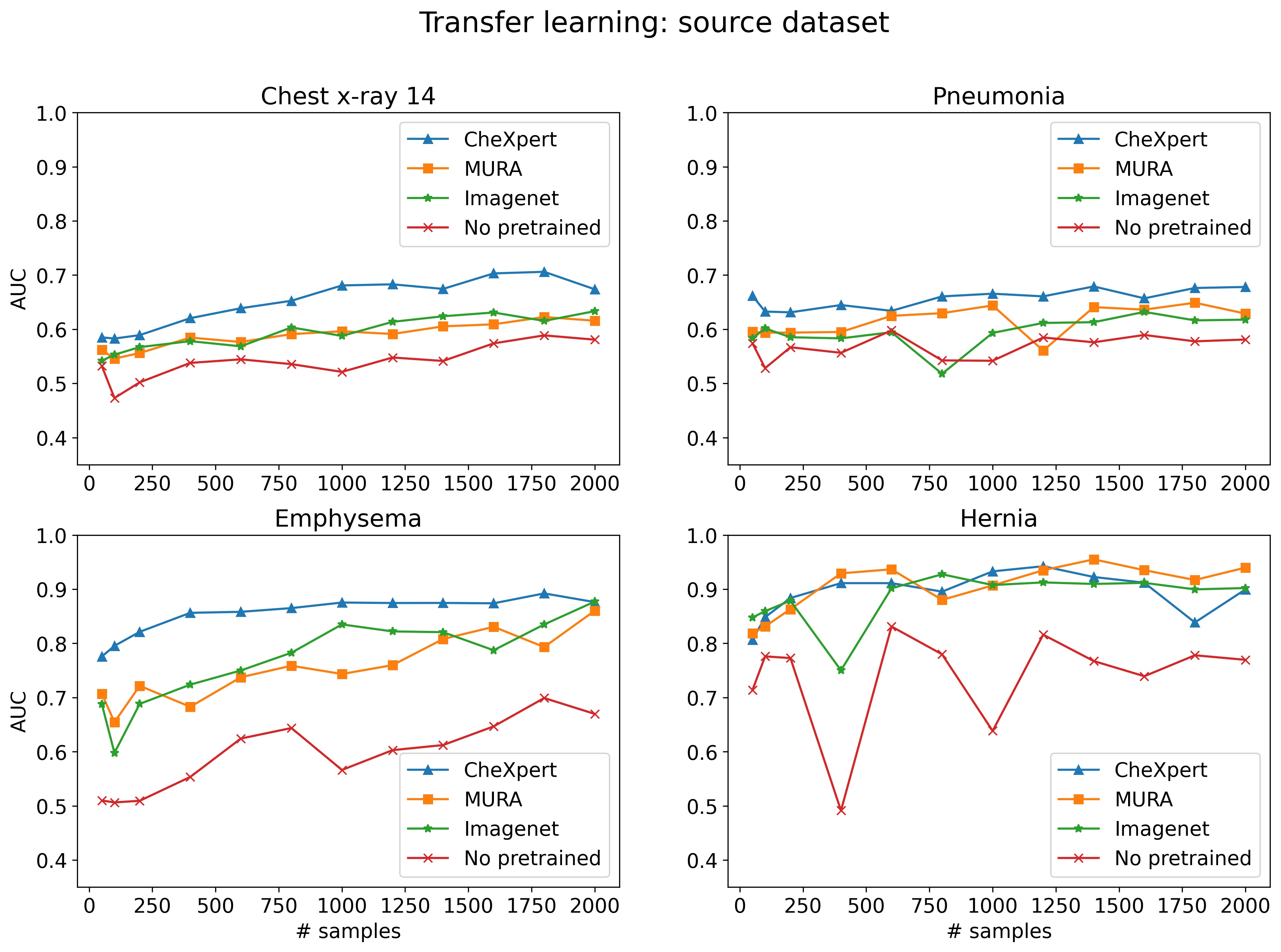}
\caption{Comparison of four and five transfer learning base datasets in terms of test AUC in a DenseNet121 over different amount of training data. Tasks (from left to right, top to bottom): Chest X-ray 14, Pneumonia, Emphysema, and Hernia.}
\label{figure: tl dataset}
\end{figure}

\section{Results}\label{sect: experiments}

In this paper the performance of deep CNN under different training and transfer learning methods has been evaluated on small training sizes for four different clinical tasks: Emphysema detection (binary classification), Pneumonia detection (binary classification), Hernia detection (binary classification) and the chest x-ray 14 task (multi-label classification). Below we describe our results:

\subsection{Deep learning models vs traditional methods}

A set of radiomic features (described in the methods section) was extracted to benchmark the performance of the CNNs. Four statistical learning algorithms were considered: logistic regression with L1 regularization, logistic regression with L2 regularization, logistic regression with elastic net (L1 and L2 regularization), and random forest. In terms of AUC, logistic regression with L1 regularization performed the best. Figure \ref{figure: all} shows the results obtained after training DenseNet121 using regular training \cite{Ng} on the Hernia and Emphysema dataset. We observe that traditional approaches can perform better than deep learning without transfer learning for very small datasets but when we use transfer learning deep learning is always better. In order to test that the deep learning with transfer learning method is significantly better than Logistic regression with Radiomic features a paired t-test was performed. This is a two-side test with the null hypothesis is that the experiments have identical mean. The tests passed with very small p-values (3.76e-08, 6.7e-08) for Emphysema and Hernia respectively.

If the deep learning models are trained with sufficient data, even with simplest approach to training, they eventually outperform traditional approaches whose performance does not see a dramatic improvement with bigger datasets due to the simplicity of the model. In medical physics and most medical imaging applications, however, it is rare to have datasets with more that 2000 points. We focus the subsequent investigation, therefore, on the performance when datasets are in the range of 50 to 2000 points.

\subsection{Comparing deep learning training methods}

Training a CNN requires identifying a strategy to select the best hyper-parameters. In our experiments we focus on the learning rate and momentum policies which are the parameters with the most significant impact on performance \cite{practicalrecommendations}. We compare regular learning rate policy and one-cycle learning rate and momentum policies for 20 epochs to determine which method converges faster and yields better results for a fixed number of epochs. Results for all four tasks are reported in Figure \ref{figure: training methods}.

We observe that one-cycle training either outperforms or preforms similarly than regular training for all tasks. Based on this observation, the remaining experiments use one-cycle training.

\begin{figure}[t]
\centering
\includegraphics[width=0.9\textwidth]{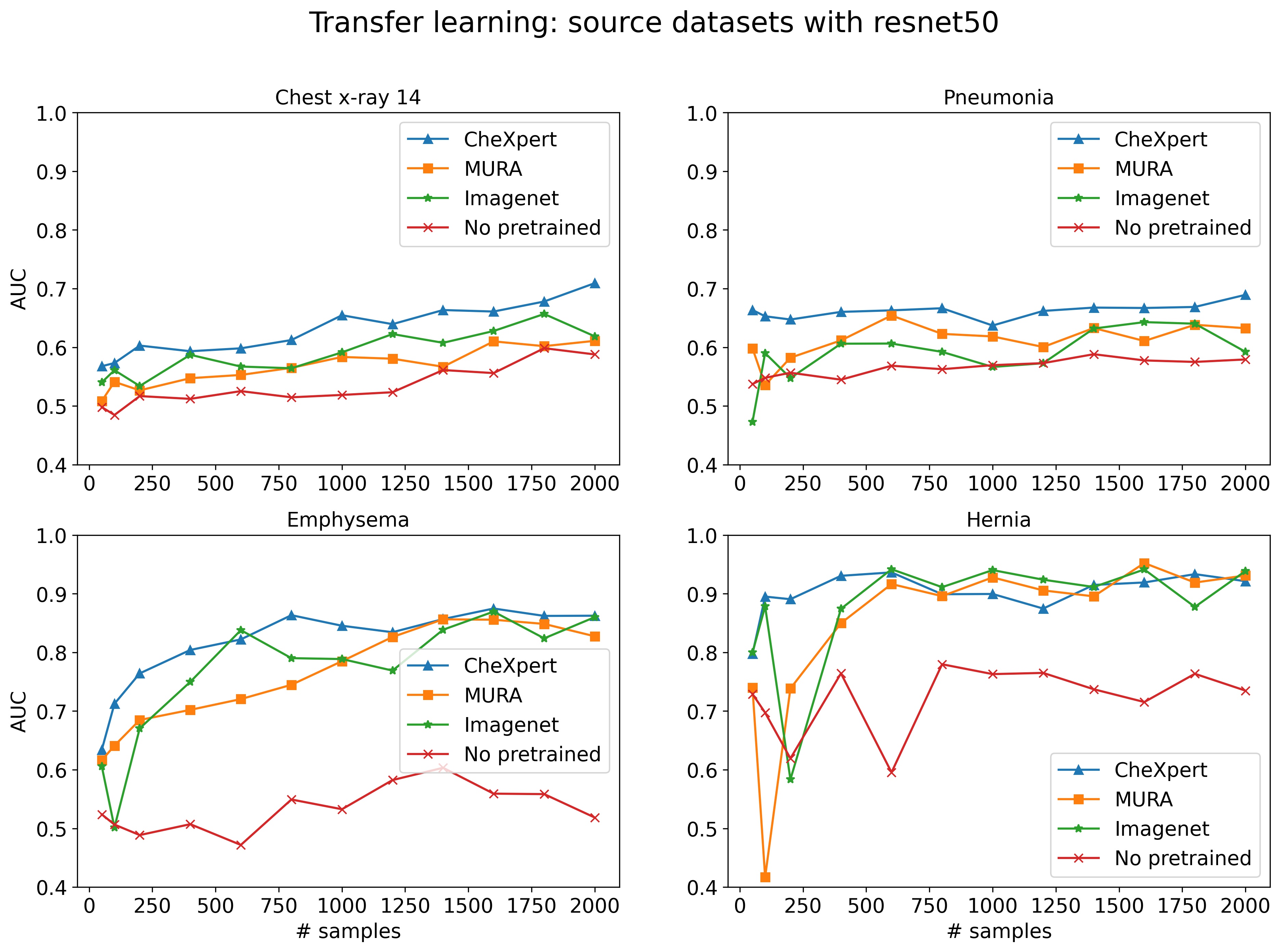}
\caption{Comparison of four different transfer learning base datasets in terms of test AUC in a Resnet50 over different amount of training data. Tasks (from left to right, top to bottom): Chest X-ray 14, Pneumonia, Emphysema, and Hernia.}
\label{figure:tl-dataset2}
\end{figure}

\begin{figure}[t]
\centering
\includegraphics[width=.5\textwidth]{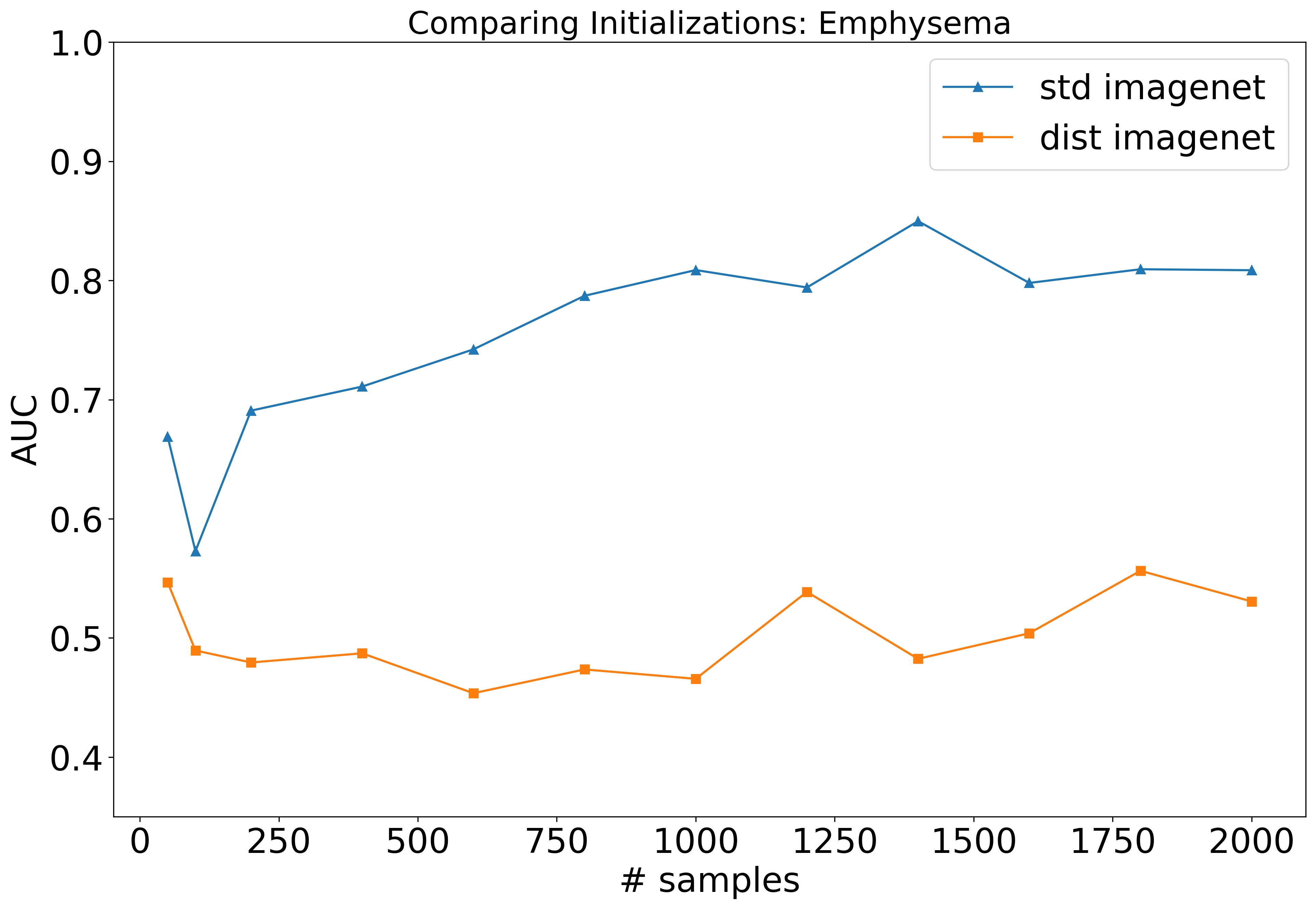}
\caption{Comparing 1st and 2nd moment transfer learning with standard transfer learning.}
\label{figure:dist}
\end{figure}

\subsection{Comparing transfer learning methods}

One-cycle learning rate and momentum policy were used as the training method to study different transfer learning strategies using networks pretrained on Imagenet. In this stage we compared three popular transfer learning methods: i) features extractor, ii) Fine-tuning of all layers with the same learning rate, and iii) gradual unfreezing with three layer groups and discriminative learning rates. The source dataset used was Imagenet, as it is almost ubiquitous used in the literature. The results are illustrated in Figure \ref{figure: tl methods}.

We observed that discriminate learning rates with gradual unfreezing and tuning all layers with the same learning rate perform almost identically. 
They both outperformed the strategy of using the pre-trained network as feature extractor for two datasets: Chest x-ray 14 and Emphysema. These results were found to be significant using a paired test with p-values of 1.01e-07 and 0.00098 respectively. For the two other datasets the three methods perform very similarly.  Although we did not find gradual unfreezing to be better than fine-tuning all layers, heuristically might still work better in certain scenarios. We also observed that transfer learning has a substantial bigger impact than the training method chose. In fact, we observe that for the Emphysema and Hernia problems we can overcome the 0.7 AUC barrier provided there are more than 500 data points available. For the subsequent experiments, therefore, we will use a one-cycle policy for training (established in the section above) and the transfer learning was done using gradual unfreezing with differential learning rates.

\subsection{Comparing source datasets in transfer learning}

The effect of the similarity of the source and target datasets are compared in terms of performance. The results are illustrated in Figure \ref{figure: tl dataset} where we observe that the boost in performance depends on the signal and difficulty of the problem, increasing from 7\% (Pneumonia) to 30\% (Emphysema). We also observe that better results are obtained when transfer learning is performed using a source dataset of exclusively the same type of images as the target dataset (e.g., frontal chest x-ray images - CheXpert). In 3 out of 4 problems (Chest x-ray 14, Emphysema, Pneumonia) the results of comparing the models pre-trained with CheXpert are significantly better than imagenet based on paired t-tests. Finally, to our surprise, the boost in performance is almost indistinguishable when it is performed using Imagenet or general x-ray images (MURA) from other body sites. In Figure \ref{figure:dist} we reproduced the same experiment using a different popular architecture: Resnet50 . We obtain very similar results which suggest that our findings do not depend on the architectures.

\subsection{Comparing initialization techniques}

It has been suggested that the value of transfer learning is not necessarily associated with the reuse of features or concepts but with having a good initialization of parameters for the network \cite{MedicalImages}. Raghu et al showed that similar performance is obtained if the parameters in the pre-trained network are changed by random numbers generated from a distribution with mean and standard deviation equal to the values obtained for each layer in the pre-trained network (preserving the first and second momentum on each layer) \cite{MedicalImages}. Such a transformation destroys the calculated features but preserves a good initialization of the parameters. Therefore, Raghul et al concluded that transfer learning might not have as deep meaning as previously thought \cite{MedicalImages}.  These results, however, were obtained for target task sizes big enough that might actually remove the value of reusing features and might not be applicable to smaller datasets \cite{MedicalImages}. Here we compare the default PyTorch initialization, standard transfer learning and preserving the first (mean) and second (standard deviation) moment of the weights to initialize the convectional layers as explained in \cite{MedicalImages}.

We observe, as shown in Figure \ref{figure:dist} that standard transfer learning from imagenet significantly outperforms preserving the 1st and 2nd moment of the layers while randomizing the parameters (by using a paired t-test with p-value 4.6e-07). Also we observe that preserving the 1st and 2nd moment of the layers while randomizing the parameters does not significantly improve the PyTorch default initialization. Therefore, for small datasets, transfer learning does play a deep role by reusing features previously calculated. Elucidating the nature of these features can bring insight in the classification problems and it will be topic of future work. We also experimented with other ways to perform the random transformation (e.g., only randomizing the convolution layers). Each of the methods yield similar results.   

\section{Discussion}

In the present article we performed an in-depth investigation of the performance of CNNs as a function of sample size, training strategy and transfer learning. We also benchmarked performance compared to using radiomic features and statistical learning algorithms. Out of bag training of deep learning models does not result in better AUC than the traditional methods for small datasets ($N < 2000$). In this case, only after careful training including transfer learning we were able to significantly outperform the radiomic method. Additionally, our results indicate that one cycle learning outperforms or is equal to regular training in all our experiments. We experimented with three transfer learning strategies: 1) features extractor, 2) fine-tune all layers with the same learning rate and 3) gradual unfreezing with three layer groups and discriminative learning rates. From these, 2 and 3 significantly outperformed 1 in two of our datasets but were similar in the two other datasets. Methods that 2 and 3 were indistinguishable from each other. These results agree with those reported by Raghu et al in larger medical datasets \cite{MedicalImages} as well as those reported by  Yosinski et al. \cite{deepNN} who point to co-adaptation as the explanation for poorer performance when networks are used as feature extractors; that is, neuron parameters in different layers evolve together in a way that is not easily discoverable if only certain parts of the network are retrained. 

Additionally, from all the strategies studied,  the biggest impact was the similarity of the source and target datasets used for transfer learning. Performing transfer learning from images trained on Imagenet (general images such as cats, dogs, etc.) or MURA (X-ray images on different parts of the body but not chest) improved results compared to scenarios when transfer learning was not used at all. Surprisingly, however, using MURA did not offer an advantage over Imagenet. More importantly, when transfer learning was performed using a source dataset that uses the same images (with different tasks), or that uses different images but of the same anatomy, performance improved significantly compared to any other methodology. This seems to indicate that for small datasets, in order to truly take advantage of transfer learning, the source images should be similar to those in the task. It further indicates that if the images are not of the same anatomy, there is little advantage in using them over Imagenet (although either strategy is better than not using transfer learning at all). These results are consistent with the hypothesis that different outcomes (or diseases) can be explained by the same concepts present in the images, as long as these images are from the same anatomy. Additionally, it also indicates that although these concepts are shared in the same part of the body, they are not as general or completely transferable to those learned with Imagenet or images from another anatomical site. 

The observations presented here are very important when models are being constructed with small datasets. If researchers keep track of multiple diseases detected for a type of image, they can use this information to obtain pretrained models better than those pretrained with Imagenet. For example, in a model tracking 13 diseases, the amount of additional (balanced) data needed to extend the model to a new disease to obtain good results is fairly minimal (recall an AUC of 0.8 was obtained with 50 points for predicting emphysema, which constitutes more than 30\% improvement over the AUC obtained without transfer learning and $14\%$ improvement over the one obtained using Imagenet. These results offer an improvement on AUC as the one obtained by Krizhevesky et al that popularized Deep Learning when they improved performance by $11\%$ over the closest competitor \cite{krizhevsky2012Imagenet}. Also note that even with the best combination for training neural networks and after a work that took a year, we could not overcome a barrier of 70\% AUC in 2 out of the four problems with fewer that 2000 training data points . This represents more than 20\% performance decrease compared to the scenario where all the data is used. Therefore, although specific training and transfer learning techniques with appropriate source datasets as those used here can provide improvement over standard methods, there is no substitution for the efficacy of big datasets \cite{sizeofdata}. Our results point to the importance of sharing datasets among our institutions, and stress the relevance of developing data-sharing platforms in medical physics  \cite{Sharedata}. This task will require a lot of political will though. This research intends to provide the science to motivate such action.

\section{Conclusions}

If enough data is available ($N > 10,000$), little to no tweaking is needed to obtain state of the art performance using standard deep learning packages today. In contrast, for small datasets, the highest boost in performance comes from a good selection of the source dataset and careful training. While transfer learning using x-rays images from other anatomical sites improves performance, we observe a similar boost by using pretrained networks from Imagenet. Having source images from the same anatomical site outperforms every other methodology. Additionally, careful training with one-cycle training and discriminative learning rates with gradual unfreezing also helps, though less than transfer learning. Finally, for small datasets, features learned in other tasks are indeed important to improve performance. These features, however, are not as general as initially thought, and performance improvement is closely linked to how similar the tasks are. There is not substitution for sufficient data and medical physicists should commit to creating big repositories for individual problems. 

\section{Acknowledgments}
Miguel Romero was partially supported by the Wicklow AI and Medical Research Initiative at the Data institute, USF. Gilmer Valdes was supported by the National Institute Of Biomedical Imaging And Bioengineering of the National Institutes of Health under Award Number K08EB026500. The content is solely the responsibility of the authors and does not necessarily represent the official views of the National Institutes of Health. The authors have no conflicts to disclose.

\textbf{References} 

\bibliography{references}
\bibliographystyle{unsrt}

Fig 8: Comparing 1st and 2nd moment transfer learning with standard transfer learning 

\end{document}